\documentclass[11pt]{article}
\usepackage{coling2020}
\usepackage{times}
\usepackage{booktabs}
\usepackage{graphicx}
\usepackage{url}
\usepackage{latexsym}
\usepackage[table,xcdraw]{xcolor}

\usepackage[utf8]{inputenc}
\usepackage{graphicx}
\usepackage{caption}
\usepackage{subcaption}
\usepackage{comment}

\title{LT@Helsinki at SemEval-2020 Task 12: \\Multilingual or language-specific BERT?}

\author{Marc Pàmies \\
  \And
  Emily Öhman \\
  \qquad\qquad\qquad\qquad\qquad University of Helsinki \\
  \qquad\qquad\qquad\qquad\qquad{\tt firstname.lastname@helsinki.fi} 
 \And  
  Kaisla Kajava \\
 \And
  Jörg Tiedemann\\
 \\}
\date{May 31, 2020}
\begin{document}
\maketitle
\begin{abstract}
This paper presents the different models submitted by the LT@Helsinki team for the SemEval 2020 Shared Task 12. Our team participated in sub-tasks A and C; titled offensive language identification and offense target identification, respectively. In both cases we used the so-called Bidirectional Encoder Representation from Transformer (BERT), a model pre-trained by Google and fine-tuned by us on the OLID and SOLID datasets. 
The results show that offensive tweet classification is one of several language-based tasks where BERT can achieve state-of-the-art results.
\end{abstract}

\section{Introduction}
\label{section1}
\blfootnote{
    %
    %
    %
    %
    %
    %
     \hspace{-0.65cm}  
     This work is licensed under a Creative Commons 
     Attribution 4.0 International License.
     License details:
     \url{http://creativecommons.org/licenses/by/4.0/}.
}

The number of social media users has reached 3.5 billion, and  an average of 6,000 tweets are generated every second \cite{10.1145/3292522.3326034} 
. With such a large volume of tweets, it seems inevitable that some use offensive language. 
A study from 2014 found that 67\% of social media users had been exposed to online hate, and 21\% had been the target of online hate \cite{oksanen2014exposure}.

The usual approach on social media sites is to forbid hate speech in the terms of service and to censor any inappropriate content detected by their algorithms, but still, companies like Facebook or Twitter have been harshly criticized for not doing enough 
\cite{del2017hate}. This criticism has forced companies to try and find accurate and scalable solutions that solve the problem of offensive language detection using automated methods. 
Workshops dealing with offensive language, such as TRAC \cite{kumar2018proceedings}, TA-COS \cite{lefever2018ta}, or ALW1 \cite{waseem2017proceedings}, as well as shared tasks like GermEval \cite{germeval2018}, TRAC-1 \cite{trac2018}, and OffensEval 2019 \cite{offenseval2019}, are becoming more and more prevalent.

The task we address here, SemEval 2020 task 12 \cite{offenseval2020}, is titled Multilingual Offensive Language Identification in Social Media and is divided into the following sub-tasks:
\begin{quote}
\begin{description}
    \item[A.] Offensive Language Identification in several languages (Arabic, Danish, English, Greek, Turkish): whether a tweet is offensive or not.
	\item[B.] Categorization of Offense Types: whether an offensive tweet is targeted or untargeted.
	\item[C.] Offense Target Identification: whether a targeted offensive tweet is directed towards an individual, a group or otherwise.
\end{description}
\end{quote}

In this paper the system created by the LT@Helsinki team for sub-tasks A and C will be described. In sub-task A we participated in all the language tracks. For sub-task C the only language available was English. We qualified as second in sub-task C and our submission for sub-task A ranked first for Danish, seventh for Greek, eighteenth for Turkish, and forty-sixth for Arabic. In all submissions we used BERT-Base models \cite{bert} fine-tuned on each dataset provided by the task organizers. We also experimented with random forest with TF-IDF and other kinds of features, but the results on the development set were not as good as with transfer learning techniques based on pre-trained language models. We discovered that, at least for this data, the language-specific model worked better than the multilingual.


\section{Background}
\label{section2}
All the data used to train our classifiers was provided by the OffensEval organizers \cite{olid}. The tweets were retrieved from the Twitter Search API and manually labeled by at least two human annotators. As a pre-processing step, they desensitized all tweets replacing usernames and website URLs by general tokens. The following datasets were used for the offensive language identification problem:

\begin{table}[htbp!]
\centering
\begin{tabular}{
>{\columncolor[HTML]{FFFFFF}}l |
>{\columncolor[HTML]{FFFFFF}}c 
>{\columncolor[HTML]{FFFFFF}}c 
>{\columncolor[HTML]{FFFFFF}}c }
\textbf{Language} & \textbf{Training} & \textbf{Test} & \multicolumn{1}{l}{\cellcolor[HTML]{FFFFFF}\textbf{Total}} \\ \toprule
Arabic  & 8,000 & 2,000 & 10,000 \\
Danish  & 2,960 & 329 & 3,289 \\
Greek   & 8,743 & 1,544 & 10,287 \\
Turkish & 31,756 & 3,528 & 35,284
\end{tabular}
\caption{Sub-task A datasets}
\label{data_a}
\end{table}

Some of the algorithms that can be found in the literature are random forest \cite{burnap2015}, logistic regression \cite{davidson2017}, and Support Vector Machine \cite{malmasi2018profanity}, as well as deep learning approaches like Convolutional Neural Networks \cite{gamback2017} or Convolutional-GRU \cite{zhang2018}. However, in 2018 deep pre-trained language models obtained state-of-the-art results in several NLP downstream tasks, text classification being one of them. In particular, Google's Bidirectional Encoder Representations from Transformers (BERT) \cite{bert} stood above the rest for being deeply bidirectional and using the novel self-attention layers from the transformer model \cite{attention}, which allows it to better understand a word’s context by looking at both its left and right neighbours. BERT provides out-of-the-box pre-trained monolingual and multilingual models that, after massive training on general corpora, can be fine-tuned with a small amount of task-specific data and still offer excellent performance. The results published in OffensEval’s previous edition \cite{offenseval2019} proved that BERT is well suited for the offensive language detection task, since it was the chosen method of most of the top teams, including the winners of sub-tasks A \cite{nuli} and C \cite{radivchev}.

For the Danish dataset \cite{offenseval20_danish_data} we used Nordic BERT\footnote{\url{https://github.com/botxo/nordic_bert}}, which is pre-trained on Danish Wikipedia texts, Danish text from Common Crawl, Danish OpenSubtitles, and text from popular Danish online forums. All in all, the training corpus consists of over 90M sentences and almost 20M unique tokens. For the other languages we used the standard BERT-Base models \cite{bert} with no further pre-training and only the provided datasets for each language, i.e. Arabic \cite{mubarak2020arabic}, Greek \cite{offenseval20_greek_data}, and Turkish \cite{offenseval20_turkish_data}.

In sub-task C, since the language was English it was possible to use the Offensive Language Identification Dataset (OLID), provided by the organizers of OffensEval 2019. Despite consisting of different sub-tasks, all of them shared the same dataset that was annotated according to a three-level hierarchical model, so that each sub-task could use as dataset a subset of the previous sub-task’s dataset. First, all tweets were labeled as either offensive (OFF) or not offensive (NOT). Then, for sub-task B, all the offensive tweets were labeled as targeted (TIN) or untargeted insults (UNT). And finally, for the last sub-task, the third level of the hierarchy labeled targeted insults based on who was the recipient of the offense: an individual (IND), a group (GRP) or a different kind of entity (OTH). To illustrate this, Figure \ref{fig:olid} displays OLID’s label distribution.

\begin{table*}[ht] \centering
	\begin{tabular}{c c c r r | r} \toprule
		{$\textbf{A}$} & {$\textbf{B}$} & {$\textbf{C}$} & {$\textbf{Training}$} & {$\textbf{Test}$} & {$\textbf{Total}$} \\ \midrule
		OFF  & TIN & IND & 2,407 & 100 & 2,507 \\
		OFF  & TIN & GRP & 1,074 & 78  & 1,152  \\
		OFF  & TIN & OTH & 395   & 35  & 430 \\
		OFF  & UNT &  -  & 524   & 27  & 551 \\
		NOT  &  -  &  -  & 8,840 & 620 & 9,460 \\
		\midrule
		\textbf{All} &  &  & 13,240 & 860 & 14,100 \\ \bottomrule
	\end{tabular}
	\caption{Distribution of label combinations in OLID.}
	\label{fig:olid}
\end{table*}

The corpus from sub-task C in OffensEval 2019 contains 4,089 English tweets, of which 3,876 originally belonged to the training set and the remaining 213 to the test set (Figure \ref{fig:olid}). We also had at our disposal over 9M English tweets provided by the OffensEval 2020 organizers \cite{rosenthal2020}. However, these were processed by unsupervised learning methods instead of human annotators, which is why on this occasion each tweet was not associated with a label but given two values: (1) the confidence that it belongs to a specific class and (2) its standard deviation.

\section{System Overview and Experimental Setup} 




\label{section4}
In this section we discuss the system we created and setup we used for subtask A and subtask C.
\subsection{Sub-task A}
For the binary classification problem (sub-task A) two baseline methods were implemented and evaluated on the training Danish dataset: random forest and BERT.

For the random forest implementation, the pre-processing steps were lower-casing all characters, removing irrelevant punctuation marks, reducing the length of characters that appear more than two consecutive times, and converting hashtags to sentences by adding white spaces before every capital letter. Tokenization was done with the TweetTokenizer tool from NLTK. The same library was also used to perform stopword removal and word stemming. Emojis were removed after storing their “sentiment score” as a feature employing the emosent Python utility package \cite{emosent}. We also used surface-level features such as the number of URL tokens and @USER mentions, the total number of characters, punctuation marks and words in each post, average word length, percentage of capital letters, and the number of abusive terms. A ratio of 10:1 was applied when splitting the dataset into training and validation sets. Since the Danish dataset was relatively small, 10-fold cross-validation was done to obtain reliable results. Otherwise the F1 scores relied too much on which samples would fall in the validation set. The random forest implementation from scikit-learn \cite{scikitlearn} was used. The optimal parameters were found with grid search.

\begin{table}[htbp]
\centering
\begin{tabular}{
>{\columncolor[HTML]{FFFFFF}}l 
>{\columncolor[HTML]{FFFFFF}}r 
>{\columncolor[HTML]{FFFFFF}}r }
\textbf{System} & \textbf{macro-F1} \\ \toprule
All NOT         & 0.465 \\
Random Forest with TFIDF             & 0.773 \\
Multilingual BERT-Base       & 0.768 \\
Nordic BERT     & 0.804                 
\end{tabular}
\caption{Development results sub-task A, Danish dataset}
\label{dev_a}
\end{table}

The other approach was to apply pre-trained BERT models. After experimenting with both the original Base version (12-layer, 768-hidden, 12-heads, 110M parameters) and the publicly available Nordic BERT \footnote{\url{https://github.com/botxo/nordic_bert}} it was clear that the second one was better suited for the task. This shows that further pre-training of BERT can significantly boost performance, especially in cases like this where there is very little data (the Danish dataset was by far the smallest of the shared task). Our final submission was generated by the Danish version of Nordic BERT fine-tuned on the OffensEval 2020 data, using a batch size of 32 for training and 16 for both validation and testing. The learning rate of Adam optimizer was set to 2e-5 and the model was trained for 4 epochs. The sequence length was set to 128 because, even though some instances are very long (some of them are not tweets but Reddit comments), an analysis of the length distribution 
showed that only 3.4\% of the training examples reached the limit after being tokenized by BertTokenizer.


Due to time constraints we were not able to experiment in depth with all languages available for sub-task A, which is why Danish was our main focus. However, seeing that BERT could perform so well with almost no pre-processing, we decided to generate results for the other three languages as well. For Turkish and Arabic, we used the BERT-base-multilingual-cased model with a maximum sequence length of 128, a batch size of 32, and a learning rate of 2e-5. On the other hand, for Greek we used the BERT-base-uncased model after lowercasing and translating the entire dataset into English. In all cases we trained the model for 4 epochs, used BertTokenizer to tokenize the tweets, and padded and truncated the sequences to make sure each data instance had the same length.

\subsection{Sub-task C}
The target identification problem (sub-task C, English) had an additional level of difficulty with respect to the first task because the dataset was highly imbalanced and composed of three classes instead of two. Thus, in this case we focused our efforts on balancing the given dataset to prevent having a high number of misclassifications from instances of the minority class (which would notably affect the resulting macro-F1 score). In order to overcome the class imbalance problem, we trained our model on all the data from 2019 and some additional instances from the non-majority classes (GRP and OTH) of the 2020 dataset. Only the 300 OTH instances and 237 GRP instances of highest confidence were added in order to slightly increase the balance of the dataset. We experimented with different thresholds to select more samples but at the end we decided to keep the value low to ensure that all tweets used for training are tagged correctly. Finally, to overcome the class imbalance problem, an over-sampling technique with replacing was applied. We simply produced copies of instances from the minority classes to end up with a totally balanced dataset of 11,628 tweets (3,876 for each class). Under-sampling was not an option because then we would be facing a data scarcity problem, and we experimented with ratios other than 1:1:1 obtaining promising results but not significantly better than the proposed approach. Another interesting approach, which was chosen by the winners of last year’s edition \cite{radivchev}, is to modify the classification thresholds (i.e. lower the thresholds for classes OTH and GRP) to get less new examples classified as the majority class.

Then, the now balanced dataset was used as input for a BERT-base-uncased model with a maximum sequence length of 128 and batch sizes of 32, 16 and 8 for training, validation and prediction respectively. The learning rate was set to 2e-5 and the training lasted 4 epochs. In this case we did not perform any pre-processing step other than lowercasing since none of the attempts (i.e. translating emojis to sentences, splitting hashtags into separate words etc.) significantly boosted performance. Moreover, we believe that for this specific task it might not be a good idea to remove @USER mentions or certain stop words since they can carry valuable information for target identification.

\subsection{Unfruitful Attempts at Performance Improvement}
\begin{table}[htbp!]
\begin{tabular}{|l|l|l|l|l|l|l|l|l|l|l|}
\hline
\cellcolor[HTML]{9B9B9B} & \textit{pos} & \textit{neg} & \textit{ang} & \textit{ant} & \textit{dis} & \textit{fea} & \textit{joy} & \textit{sad} & \textit{sur} & \textit{tru} \\ \hline
\textbf{NOT}             & 51.476       & 79.136       & 30.028       & 17.623       & 29.685       & 47.055       & 16.588       & 25.843       & 10.753       & 26.785       \\ \hline
\textbf{OFF}             & 68.640       & 110.350      & 41.481       & 23.474       & 37.543       & 58.420       & 22.374       & 46.046       & 9.190        & 26.838       \\ \hline
\end{tabular}
\caption{Emotion word distribution in offensive and non-offensive messages.}
\label{table:emo}
\end{table}

Offensive tweets and posts had statistically significant amounts of emotion-laden words, including positive ones. We arrived at these results by augmenting the data (see Ohman \shortcite{ohman2016challenges}) using the NRC Emotion Lexicon \cite{Mohammad13} for the languages in question and then comparing the normalized word counts per offensive post. We did not take context or negation into account. Unfortunately, we were unable to meaningfully utilize this information to improve the performance of our system. Even though offensive tweets were more likely to include emotion words and more of them, a significant number of offensive tweets contained no emotion words and many non-offensive tweets did contain them. Nonetheless, the results were quite interesting so we wanted to share them here in the hopes that they might be useful to someone else, perhaps for the OffensEval 2021 tasks. The numbers represent normalized word counts for words classified as containing a specific sentiment or emotion for data from the Danish dataset.

\section{Results}


\label{section5}
\begin{figure}[h]%
    \centering
    
    \includegraphics[scale=0.42]{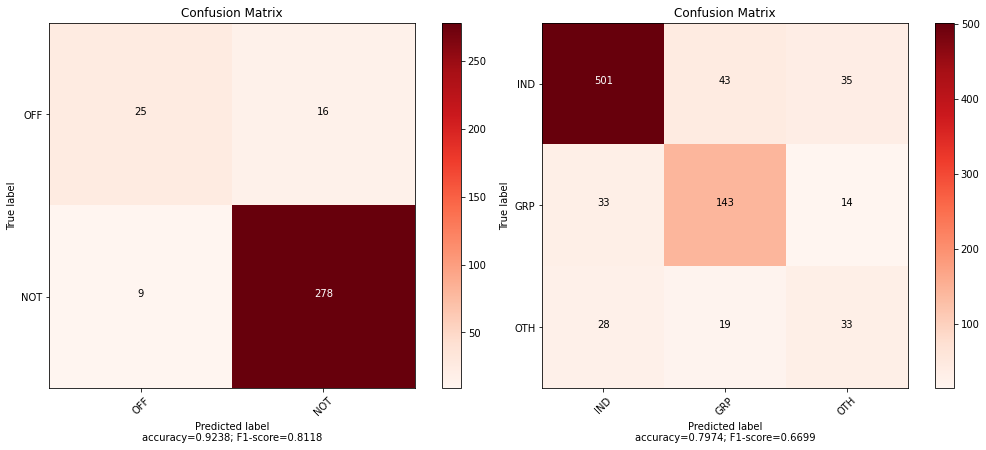}
    \caption{Official results of the LT@Helsinki Team. \\Left: Danish, sub-task A, Right: Sub-task C (English)}%
    \label{fig:cm}%
\end{figure}

For Danish, using Nordic BERT for the final submission, we obtained an accuracy of 92.38\% and F1-score of 81.18\% (Figure \ref{fig:cm} a). The confusion matrix shows that only 9 out of 34 offensive tweets (OFF) were misclassified, and 16 out of 294 not offensive tweets (NOT) were wrongly classified as offensive. Regarding the other languages, for Greek we obtained an F1-score of 82.6\%, 77.2\% for Turkish, and 73.1\% for Arabic. 

In sub-task C our submission ranked second with an accuracy of 79.74\% and F1-score of 66.99\%. The relatively low F1-score is due to the high numbers of misclassifications of the minority class (only 33 OTH instances were correctly classified out of 82) (see Figure \ref{fig:cm} b). Despite our efforts in balancing the dataset, it seems like the skewed class distribution inevitably added bias to the model.

It is impressive how good BERT performed overall even though we applied very few pre-processing steps to the data. Moreover, the results obtained on previously unseen data indicate that none of our models were heavily overfitted.


\section{Conclusions}
\label{section6}

Our scores on evaluation data show that models pre-trained on general corpora can obtain competitive results when there is very little data available, as was the case of the Danish sub-task where we obtained the best results. BERT performed well on both tasks, but there is still room for improvement. In the future, we intend to experiment with those languages that we could not focus on this time. It should be noted that multilingual BERT works best with languages similar to English \cite{pires2019multilingual} so it is very likely that the other languages would have benefited from the use of language-specific models even more than Danish as Danish is the language most similar to English in comparison to Greek, Turkish, and Arabic. 

A more thorough comparison between multilingual and language-specific BERT would yield more definitive answers as to whether language-specific is always best or whether that is language-dependent. Further examination and use of emotion-laden words in offensive texts could also help with the detection of offensive texts. Examining other baseline methods and combining them into an ensemble model with majority voting is another approach to consider for future work. With regards to the class imbalance problem, other techniques such as adjusting the classification thresholds or enriching the dataset with back and forth machine translation of the minority class could prove to be useful.

\bibliographystyle{acl}
\bibliography{offenseval}

\begin{thebibliography}{}

\bibitem[\protect\citename{Burnap and Williams}2015]{burnap2015}
Pete Burnap and Matthew~L Williams.
\newblock 2015.
\newblock Cyber hate speech on twitter: An application of machine
  classification and statistical modeling for policy and decision making.
\newblock {\em Policy \& Internet}, 7(2):223--242.

\bibitem[\protect\citename{\c{C}\"{o}ltekin}2020]{offenseval20_turkish_data}
\c{C}a\u{g}r{\i} \c{C}\"{o}ltekin.
\newblock 2020.
\newblock {A Corpus of Turkish Offensive Language on Social Media}.
\newblock In {\em Proceedings of the 12th International Conference on Language
  Resources and Evaluation}. ELRA.

\bibitem[\protect\citename{Davidson \bgroup et al.\egroup }2017]{davidson2017}
Thomas Davidson, Dana Warmsley, Michael~W. Macy, and Ingmar Weber.
\newblock 2017.
\newblock Automated hate speech detection and the problem of offensive
  language.
\newblock In {\em ICWSM}.

\bibitem[\protect\citename{Del~Vigna \bgroup et al.\egroup }2017]{del2017hate}
Fabio Del~Vigna, Andrea Cimino, Felice Dell’Orletta, Marinella Petrocchi, and
  Maurizio Tesconi.
\newblock 2017.
\newblock Hate me, hate me not: Hate speech detection on facebook.
\newblock In {\em Proceedings of the First Italian Conference on Cybersecurity
  (ITASEC17)}, pages 86--95.

\bibitem[\protect\citename{Devlin \bgroup et al.\egroup }2019]{bert}
Jacob Devlin, Ming-Wei Chang, Kenton Lee, and Kristina Toutanova.
\newblock 2019.
\newblock Bert: Pre-training of deep bidirectional transformers for language
  understanding.
\newblock In {\em NAACL-HLT}.

\bibitem[\protect\citename{Gamb{\"a}ck and Sikdar}2017]{gamback2017}
Bj{\"o}rn Gamb{\"a}ck and Utpal~Kumar Sikdar.
\newblock 2017.
\newblock Using convolutional neural networks to classify hate-speech.
\newblock In {\em Proceedings of the first workshop on abusive language
  online}, pages 85--90.

\bibitem[\protect\citename{Kumar \bgroup et al.\egroup }2018a]{trac2018}
Ritesh Kumar, Atul~Kr. Ojha, Shervin Malmasi, and Marcos Zampieri.
\newblock 2018a.
\newblock Benchmarking aggression identification in social media.
\newblock In {\em TRAC@COLING 2018}.

\bibitem[\protect\citename{Kumar \bgroup et al.\egroup
  }2018b]{kumar2018proceedings}
Ritesh Kumar, Atul~Kr Ojha, Marcos Zampieri, and Shervin Malmasi.
\newblock 2018b.
\newblock Proceedings of the first workshop on trolling, aggression and
  cyberbullying (trac-2018).
\newblock In {\em Proceedings of the First Workshop on Trolling, Aggression and
  Cyberbullying (TRAC-2018)}.

\bibitem[\protect\citename{Lefever \bgroup et al.\egroup }2018]{lefever2018ta}
Els Lefever, Bart Desmet, and Guy De~Pauw.
\newblock 2018.
\newblock Ta-cos 2018: 2nd workshop on text analytics for cybersecurity and
  online safety: Proceedings.
\newblock In {\em TA-COS 2018--2nd Workshop on Text Analytics for Cybersecurity
  and Online Safety, collocated with LREC 2018, 11th edition of the Language
  Resources and Evaluation Conference}. European Language Resources Association
  (ELRA).

\bibitem[\protect\citename{Liu \bgroup et al.\egroup }2019]{nuli}
Ping Liu, Wen Li, and Liang Zou.
\newblock 2019.
\newblock Nuli at semeval-2019 task 6: transfer learning for offensive language
  detection using bidirectional transformers.
\newblock In {\em Proceedings of the 13th International Workshop on Semantic
  Evaluation}, pages 87--91.

\bibitem[\protect\citename{Malmasi and Zampieri}2018]{malmasi2018profanity}
Shervin Malmasi and Marcos Zampieri.
\newblock 2018.
\newblock Challenges in discriminating profanity from hate speech.
\newblock {\em Journal of Experimental \& Theoretical Artificial Intelligence},
  30(2):187--202.

\bibitem[\protect\citename{Mathew \bgroup et al.\egroup
  }2019]{10.1145/3292522.3326034}
Binny Mathew, Ritam Dutt, Pawan Goyal, and Animesh Mukherjee.
\newblock 2019.
\newblock Spread of hate speech in online social media.
\newblock In {\em Proceedings of the 10th ACM Conference on Web Science},
  WebSci ’19, page 173–182, New York, NY, USA. Association for Computing
  Machinery.

\bibitem[\protect\citename{Mohammad and Turney}2013]{Mohammad13}
Saif~M. Mohammad and Peter~D. Turney.
\newblock 2013.
\newblock Crowdsourcing a word-emotion association lexicon.
\newblock {\em Computational Intelligence}, 29(3):436--465.

\bibitem[\protect\citename{Mubarak \bgroup et al.\egroup
  }2020]{mubarak2020arabic}
Hamdy Mubarak, Ammar Rashed, Kareem Darwish, Younes Samih, and Ahmed Abdelali.
\newblock 2020.
\newblock Arabic offensive language on twitter: Analysis and experiments.
\newblock {\em arXiv preprint arXiv:2004.02192}.

\bibitem[\protect\citename{Nikolov and Radivchev}2019]{radivchev}
Alex Nikolov and Victor Radivchev.
\newblock 2019.
\newblock Nikolov-radivchev at semeval-2019 task 6: Offensive tweet
  classification with bert and ensembles.
\newblock In {\em Proceedings of the 13th International Workshop on Semantic
  Evaluation}, pages 691--695.

\bibitem[\protect\citename{Novak \bgroup et al.\egroup }2015]{emosent}
Petra~Kralj Novak, Jasmina Smailovi{\'c}, Borut Sluban, and Igor Mozeti{\v{c}}.
\newblock 2015.
\newblock Sentiment of emojis.
\newblock {\em PloS one}, 10(12):e0144296.

\bibitem[\protect\citename{Oksanen \bgroup et al.\egroup
  }2014]{oksanen2014exposure}
Atte Oksanen, James Hawdon, Emma Holkeri, Matti N{\"a}si, and Pekka
  R{\"a}s{\"a}nen.
\newblock 2014.
\newblock Exposure to online hate among young social media users.
\newblock {\em Sociological studies of children \& youth}, 18(1):253--273.

\bibitem[\protect\citename{Pedregosa \bgroup et al.\egroup }2011]{scikitlearn}
Fabian Pedregosa, Ga{\"e}l Varoquaux, Alexandre Gramfort, Vincent Michel,
  Bertrand Thirion, Olivier Grisel, Mathieu Blondel, Peter Prettenhofer, Ron
  Weiss, Vincent Dubourg, et~al.
\newblock 2011.
\newblock Scikit-learn: Machine learning in python.
\newblock {\em Journal of machine learning research}, 12(Oct):2825--2830.

\bibitem[\protect\citename{Pires \bgroup et al.\egroup
  }2019]{pires2019multilingual}
Telmo Pires, Eva Schlinger, and Dan Garrette.
\newblock 2019.
\newblock How multilingual is multilingual bert?
\newblock {\em arXiv preprint arXiv:1906.01502}.

\bibitem[\protect\citename{Pitenis \bgroup et al.\egroup
  }2020]{offenseval20_greek_data}
Zeses Pitenis, Marcos Zampieri, and Tharindu Ranasinghe.
\newblock 2020.
\newblock {Offensive Language Identification in Greek}.
\newblock In {\em Proceedings of the 12th Language Resources and Evaluation
  Conference}. ELRA.

\bibitem[\protect\citename{Rosenthal \bgroup et al.\egroup
  }2020]{rosenthal2020}
Sara Rosenthal, Pepa Atanasova, Georgi Karadzhov, Marcos Zampieri, and Preslav
  Nakov.
\newblock 2020.
\newblock {A Large-Scale Semi-Supervised Dataset for Offensive Language
  Identification}.
\newblock In {\em arxiv}.

\bibitem[\protect\citename{Sigurbergsson and
  Derczynski}2019]{offenseval20_danish_data}
Gudbjartur~Ingi Sigurbergsson and Leon Derczynski.
\newblock 2019.
\newblock Offensive language and hate speech detection for danish.
\newblock {\em arXiv preprint arXiv:1908.04531}.

\bibitem[\protect\citename{Vaswani \bgroup et al.\egroup }2017]{attention}
Ashish Vaswani, Noam Shazeer, Niki Parmar, Jakob Uszkoreit, Llion Jones,
  Aidan~N Gomez, {\L}ukasz Kaiser, and Illia Polosukhin.
\newblock 2017.
\newblock Attention is all you need.
\newblock In {\em Advances in neural information processing systems}, pages
  5998--6008.

\bibitem[\protect\citename{Waseem \bgroup et al.\egroup
  }2017]{waseem2017proceedings}
Zeerak Waseem, Wendy Hui~Kyong Chung, Dirk Hovy, and Joel Tetreault.
\newblock 2017.
\newblock Proceedings of the first workshop on abusive language online.
\newblock In {\em Proceedings of the First Workshop on Abusive Language
  Online}.

\bibitem[\protect\citename{Wiegand \bgroup et al.\egroup }2018]{germeval2018}
Michael Wiegand, Melanie Siegel, and Josef Ruppenhofer.
\newblock 2018.
\newblock {Overview of the GermEval 2018 Shared Task on the Identification of
  Offensive Language}.
\newblock In {\em Proceedings of GermEval}.

\bibitem[\protect\citename{Zampieri \bgroup et al.\egroup }2019a]{olid}
Marcos Zampieri, Shervin Malmasi, Preslav Nakov, Sara Rosenthal, Noura Farra,
  and Ritesh Kumar.
\newblock 2019a.
\newblock Predicting the type and target of offensive posts in social media.
\newblock {\em arXiv preprint arXiv:1902.09666}.

\bibitem[\protect\citename{Zampieri \bgroup et al.\egroup
  }2019b]{offenseval2019}
Marcos Zampieri, Shervin Malmasi, Preslav Nakov, Sara Rosenthal, Noura Farra,
  and Ritesh Kumar.
\newblock 2019b.
\newblock {SemEval-2019 Task 6: Identifying and Categorizing Offensive Language
  in Social Media (OffensEval)}.
\newblock In {\em Proceedings of The 13th International Workshop on Semantic
  Evaluation (SemEval)}.

\bibitem[\protect\citename{Zampieri \bgroup et al.\egroup
  }2020]{offenseval2020}
Marcos Zampieri, Preslav Nakov, Sara Rosenthal, Pepa Atanasova, Georgi
  Karadzhov, Hamdy Mubarak, Leon Derczynski, Zeses Pitenis, and
  \c{C}a\u{g}r{\i} \c{C}\"{o}ltekin.
\newblock 2020.
\newblock {SemEval-2020 Task 12: Multilingual Offensive Language Identification
  in Social Media (OffensEval 2020)}.
\newblock In {\em Proceedings of SemEval}.

\bibitem[\protect\citename{Zhang \bgroup et al.\egroup }2018]{zhang2018}
Ziqi Zhang, David Robinson, and Jonathan Tepper.
\newblock 2018.
\newblock Detecting hate speech on twitter using a convolution-gru based deep
  neural network.
\newblock In {\em European semantic web conference}, pages 745--760. Springer.

\bibitem[\protect\citename{{Ö}hman \bgroup et al.\egroup
  }2016]{ohman2016challenges}
Emily {Ö}hman, Timo Honkela, and Jörg Tiedemann.
\newblock 2016.
\newblock The challenges of multi-dimensional sentiment analysis across
  languages.
\newblock In {\em Proceedings of the Workshop on Computational Modeling of
  People’s Opinions, Personality, and Emotions in Social Media (PEOPLES)},
  pages 138--142.

\end{thebibliography}
\end{document}